\documentclass[letterpaper, 10 pt, conference]{ieeeconf}  

\IEEEoverridecommandlockouts                              

\usepackage{graphics} 
\usepackage{epsfig} 
\usepackage{times} 
\usepackage{amsmath} 
\usepackage{amssymb}  

\newcommand{\be}{\begin{equation}}
\newcommand{\ee}{\end{equation}}

\DeclareMathOperator*{\argmin}{arg\,min}

\title{\LARGE \bf
Automated Map Reading: Image Based Localisation in\\[0.5ex] 2-D Maps Using Binary Semantic Descriptors
}

\author{Pilailuck Panphattarasap and Andrew Calway
\thanks{The authors are with Department of Computer Science, University of Bristol, UK        {\tt\small \{pp12907,andrew.calway\}@bristol.ac.uk}}%
}

\begin{document}

\maketitle
\thispagestyle{empty}
\pagestyle{empty}

\begin{abstract}

We describe a novel approach to image based localisation in urban environments using semantic matching between images and a 2-D map. It contrasts with the vast majority of existing approaches which use image to image database matching. We use highly compact binary descriptors to represent semantic features at locations, significantly increasing scalability compared with existing methods and having the potential for greater invariance to variable imaging conditions. The approach is also more akin to human map reading, making it more suited to human-system interaction. The binary descriptors indicate the presence or not of semantic features relating to buildings and road junctions in discrete viewing directions. We use CNN classifiers to detect the features in images and match descriptor estimates with a database of location tagged descriptors derived from the 2-D  map. In isolation, the descriptors are not sufficiently discriminative, but when concatenated sequentially along a route, their combination becomes highly distinctive and allows localisation even when using non-perfect classifiers. Performance is further improved by taking into account left or right turns over a route. Experimental results obtained using Google StreetView and OpenStreetMap data show that the approach has considerable potential, achieving localisation accuracy of around 85\% using routes corresponding to approximately 200 meters.

\end{abstract}

\section{INTRODUCTION}

Image based localisation and place recognition have been looked at extensively as an alternative to infrastructure dependent sensing such as GPS, especially when operating in urban environments. The vast majority of systems adopt an {\em image to image database matching} approach, in which environment images are matched to a database of location tagged images \cite{Lowry-TOR-2015}. Although these have demonstrated impressive performance, they are also limited in three key respects. The first is scalability - localisation is dependent on having a very large database of images or image features and thus scaling to very large areas is problematic. The second relates to invariance - matching is impacted significantly by variable imaging conditions and so maintaining performance at all times over extended periods is challenging. Finally, such schemes do not align well with how it is believed  that humans perceive and undertake location-based activities, which are thought to be based on some form of 2-D map representation \cite{Lynch1960,O'Keefe1978,Tversky1993}, and thus these approaches do not lend themselves naturally to human-system interaction.

Motivated by the above, we consider an alternative approach using {\em image to 2-D map matching}, in which we link images to semantic features on a 2-D map of an environment to give localisation. We therefore move away from matching images and instead match semantic information. This is akin to human map reading, in which a person relates the surrounding visual appearance of an environment to the semantic information they can perceive on a map, such as buildings, road layout, etc. This renders the approach better suited to human-system interaction. Moreover, the abstraction and compression provided by semantic description also gives potential for significant gains in scalability - our semantic descriptors are many orders of magnitude smaller than images or sets of image features - and improved invariance to variable imaging conditions, since via training, the detection of semantic features in images can be made less dependent on specific appearance. 

\begin{figure}[t]
\begin{center}
\includegraphics[width=0.8\columnwidth]{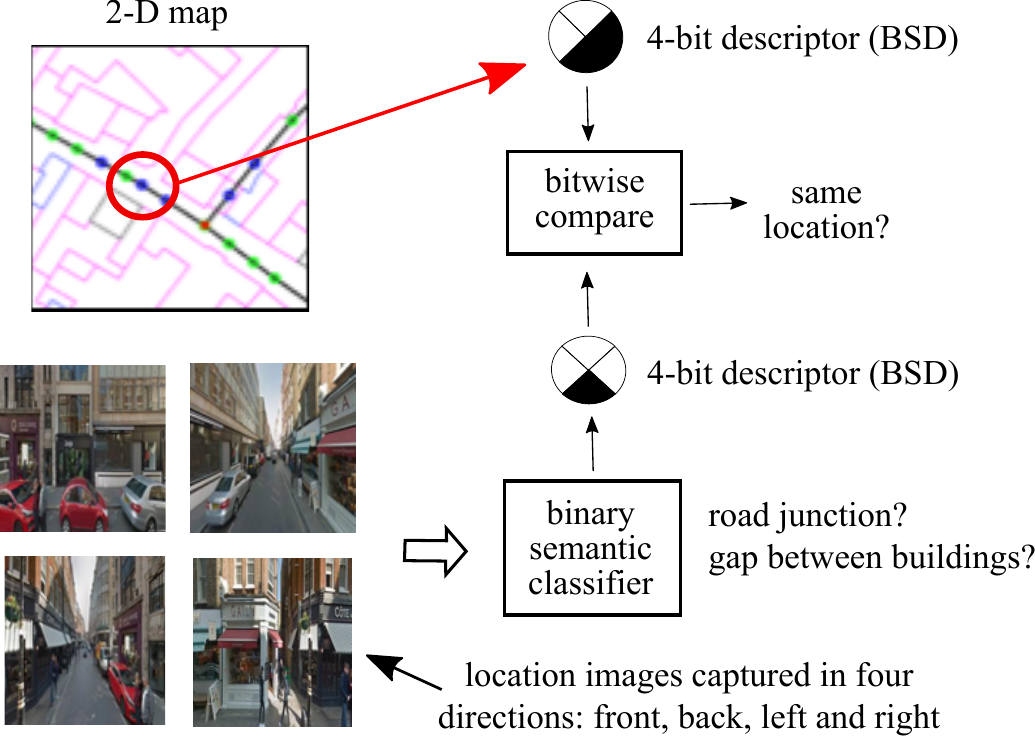}
\end{center}
\vspace*{-1ex}
\caption{Binary semantic descriptors (BSDs). 4-bit binary descriptors are used to represent locations indicating the presence or not of semantic features in 4 directions (front/back facing - junctions; left/right facing - gaps between buildings). These are derived from a 2-D map and compared bitwise with descriptors estimated via classifiers from images captured in the same directions to establish localisation w.r.t the map. On their own the descriptors are not sufficiently distinctive, but when combined sequentially along routes as shown in Figure \ref{fig:overview}, then localisation becomes possible.}
\label{fig:bsd}
\vspace*{-3ex}
\end{figure}

In this paper we present preliminary investigations into the approach. Our central idea is to characterise locations by a small number of semantic features relating to road junctions, buildings, etc, and then represent each location by a binary semantic descriptor (BSD), with each bit indicating the presence or not of a given feature in a given viewing direction. This gives a very compact representation (we use 4-bit descriptors in this work) and so increases scalability. We design classifiers to recognise the features in images, allowing us to estimate the descriptors and hence in principle recognise locations by comparison with a database of location tagged descriptors derived directly from the 2-D map. The approach is illustrated in Figure \ref{fig:bsd}. 

\begin{figure}[t]
\begin{center}
\vspace{2ex}
\includegraphics[width=\columnwidth]{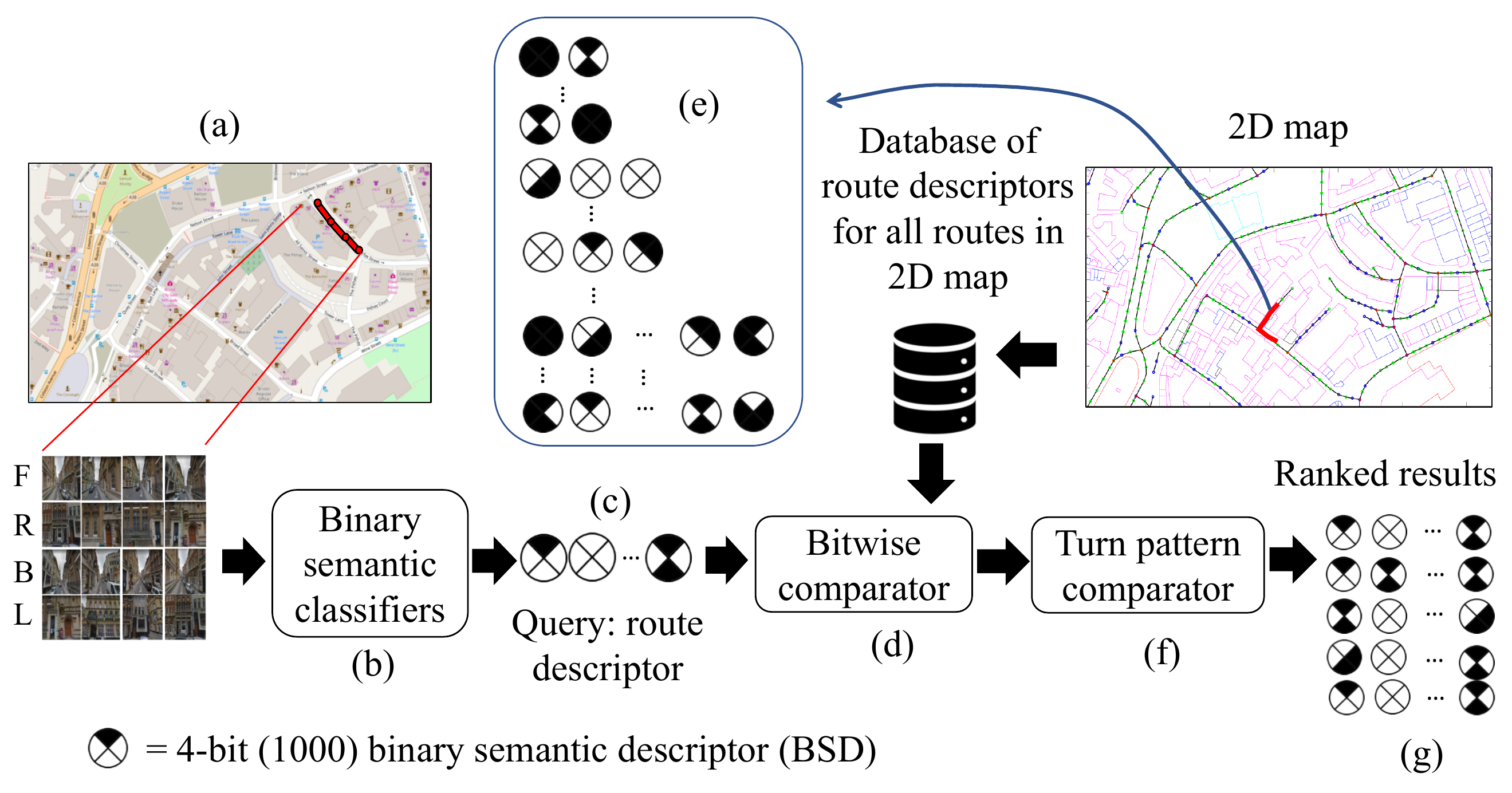}
\end{center}
\vspace*{-3ex}
\caption{Route based localisation. (a) Images captured in four directions (front, back, left and right facing) at locations along a route are converted to BSDs using binary classifiers (b) and concatenated to produce route descriptors (c). These are compared bit-wise (d) with a database of ground-truth BSDs (e) derived from the 2-D map to determine the closest matching route. Routes are then compared in terms of their turn patterns (f) to give a final ranking of possible locations of the images w.r.t the 2-D map (g).}
\label{fig:overview}
\vspace*{-3ex}
\end{figure}

However, due to their simplicity, the above descriptors are not sufficiently discriminative on their own; there are many locations having the same descriptors and when coupled with non-perfect classifiers, localisation is not possible. Nevertheless, when the descriptors are concatenated sequentially, then the resulting {\em route descriptors}  do become highly distinctive, to the extent that localisation is possible despite non-perfect classifiers. In essence, the {\em pattern of semantic features} observed along a route become unique providing the route is sufficiently long (in the experiments reported below we achieved localisation after approximately 200 meters). Moreover, when the direction of travel between locations along a route is also taken into account, e.g. left and right turns, performance is further improved. This routes based approach is illustrated in Figure \ref{fig:overview}. Note that it is feasible because of the compact nature of the map representation, i.e. a small number of bits per location, and is something that would be difficult to achieve using the comparatively large representations used in image to image database matching. 

In this paper, we present an implementation using Google StreetView (GSV) and OpenStreetMap\footnote{www.openstreetmap.org} (OSM) data, with the latter providing vector maps and the former giving 360 degree images at regular locations along roads. We used road junctions and gaps between buildings as our semantic features, assuming the former to be present or not in front and back facing views, and the latter to be present or not in left and right facing views. This gives us 4-bit descriptors for each location. We trained convolutional neural network (CNN) classifiers to recognise the features in images, achieving accuracy of around 75\%. In experiments on an area of around $2.5$ $\mbox{km}^2$, we achieved localisation accuracy in excess of 85\% when using routes consisting of 20 or more locations, corresponding to distances of approximately 200 meters. Although initial localisation is delayed as the route evolves, once bootstrapped to the correct location, the method successfully tracks the route at the same rate as location images are captured and achieves this using a significantly smaller database than required in  image to image database matching. The results suggest that the method has considerable potential.

\section{RELATED WORK}

Approaches to image based localisation and place recognition have almost exclusively focused on image to image database matching, in which environments are represented by sets of location tagged images or image features \cite{Lowry-TOR-2015}. The key concerns in such methods are the invariance of representations to changes in viewpoint and changes in appearance caused by different lighting and weather conditions. For example, the FAB-MAP algorithms \cite{Cummins2010} use image features with a degree of viewpoint invariance to give large-scale matching over long routes of up to 1000 km, whilst other methods have sought to deal with changing appearance either through invariant representations \cite{Milford2012}, storing multiple representations \cite{Biber2009} or learning models of appearance change \cite{Lowry2014}. More recent work has looked at leveraging the power of deep learning methods to gain improved  matching \cite{Suenderhauf2015b,Panphattarasap2016}. However, in all cases, large scale localisation requires large scale memory requirements, in the order of hundreds of gigabytes \cite{Cummins2010}. 

In contrast, although there is a body of work which has looked at using computer vision to extract navigation features from paper maps, see e.g. the survey in \cite{chiang2014}, there has been very little work on linking maps to images for localisation as described in this paper. There has been some work on utilising semantic information in the form of identifying key landmarks and objects in images, such as buildings, traffic lights, bollards, etc, and using these to represent locations, see e.g. \cite{Frampton2013,Mousavian2015}. These approaches have the potential to provide good invariance, including with ultra-wide viewpoint changes \cite{Frampton2013}, and reduced representational size, but to date they have been limited in scale and not linked to map information. Closer in spirit to our work in terms of alignment with human wayfinding is the PhotoMap application described in \cite{Cheverst2008} and \cite{Schoning2009}. Images of `You are here' public maps are geo-referenced with online maps by hand to provide specialised local data alongside navigation information on mobile devices, recognising the value of pictorial map data for human spatial cognition.

There has been some recent work on estimating 6-D camera pose using a combination of GPS, images and map data in urban environments as described in \cite{Cham2010,Mousavian2016,Arth2015,Armagan2017a,Armagan2017b}. The methods described use building edges and planar facades extracted from images to align with 2-D and 2.5-D maps geo-localised using GPS and so give improved estimates of camera position and orientation. However, these methods focus on obtaining precise metric estimates of camera pose for applications such Augmented Reality \cite{Arth2015} based on clear views of building facades. As such they would be difficult to extend to general localisation.

The closest work to that presented here is that described by Seff and Xiao \cite{Seff2016}. In a similar manner to our detectors described below, they use a CNN approach to recognise semantic features in images of urban settings, such as junctions, number of lanes, drivability, bike lanes, one-way vs two-way, etc. The network training is based on ground-truth features obtained from OSM and images from GSV, in much the same manner as in our approach. However, their focus is on using the outputs of the classifier to validate map locations provided by GPS for self-driving car applications, rather than for general localisation. In addition, they consider locations in isolation, in contrast to our use of route  information. 

Given the above and to the best of our knowledge, we therefore believe that the approach presented here is the first of its kind in terms of systematically linking 2-D map data with images for position localisation over large  areas.

\section{OVERVIEW}

The main components of the approach are illustrated in Figure \ref{fig:overview}. From a 2-D vector map, i.e. OSM, we generate binary semantic descriptors (BSDs) for locations spaced at regular intervals along roads in an urban environment. Each descriptor consists of 4 bits, with each bit indicating the presence or not of a semantic feature in a given viewing direction. We used four directions - front, back, left and right facing - and two feature types - junctions and gaps between buildings. The latter were chosen since they are easily identified in the vector map and as described below, they can be reliably detected in images using trained classifiers. 

A database of location tagged route descriptors is then created by computing all possible routes within the area of interest up to a certain length in terms of the number of adjacent locations and then concatenating the set of associated BSDs as indicated in Figure \ref{fig:overview}d, where the circular discs represent the BSDs and the black/white segments indicate individual bits. Note that each route descriptor is then of length $4N_r$ bits, where $N_r$ is the number of locations in the route. Thus, although the number of possible routes can be very large, the route database has a small memory footprint. For example, in the experiments described below, for an area of approximately $2.5$ $\mbox{km}^2$, the number of possible routes containing 40 locations (each approximately 400 meters long represented by a 160-bit route descriptor) is just under $40\times 10^6$. The route descriptor database is then around 800 MB in raw form, i.e. prior to any compression, which would be possible due to significant overlap between routes. This contrasts, for example, with the 177 GB reported in \cite{Cummins2010} required for image features to represent a single 1000 km route, i.e. equivalent to 71 MB for a single 400 meter route.

Localisation w.r.t the map then proceeds as follows. Images in the four viewing directions are captured at a location, i.e. within GSV in our case. Each image is then fed to a binary classifier, which detects the presence or not of a semantic feature, i.e. a junction for the front and back facing views and a gap between buildings for the left and right facing views. This gives a 4-bit BSD as illustrated in Figure \ref{fig:bsd}, with each bit indicating the presence or not of the feature in each viewing direction. 

The above BSD could be compared with those for all locations in the 2-D map to give localisation, but as noted earlier, their simplicity means that they are not sufficiently distinctive, with many locations having the same descriptor. Instead, as shown in in Figure \ref{fig:overview}a-c, we concatenate BSDs as the 'user' moves along a route in the environment, capturing images and generating descriptors at regular intervals, creating a route descriptor. In our case, we have a virtual user moving in GSV and generate BSDs at each successive GSV location (approximately every 10 meters). At each location, the current route descriptor is then used to query the database, with Hamming distances used to provide a ranked list of likely locations, as illustrated in Figure \ref{fig:overview}d-g.

To add further discrimination, we also compare the {\em turn patterns} - the position of left or right turns in a route - associated with the query and database routes, requiring that these are identical for a valid match. The motivation here is that direction changes of, for example, an autonomous vehicle can be detected reliably and hence can be used to eliminate spurious matches between route descriptors. The database route having the lowest Hamming distance w.r.t the query route and also the same turn pattern then provides the location estimate. 

In the following sections, we provide details of the BSD generation, the design and training of the binary classifiers, the generation and comparison of the turn patterns and a probabilistic interpretation of the approach. Section \ref{sec:expts} provides details of the GSV/OSM experiments and results and we conclude with a brief discussion of future work.

\section{BINARY SEMANTIC DESCRIPTORS \label{sec:bsd}}

\begin{figure}[t]
\vspace*{1ex}
\begin{center}
\includegraphics[width=0.8\columnwidth]{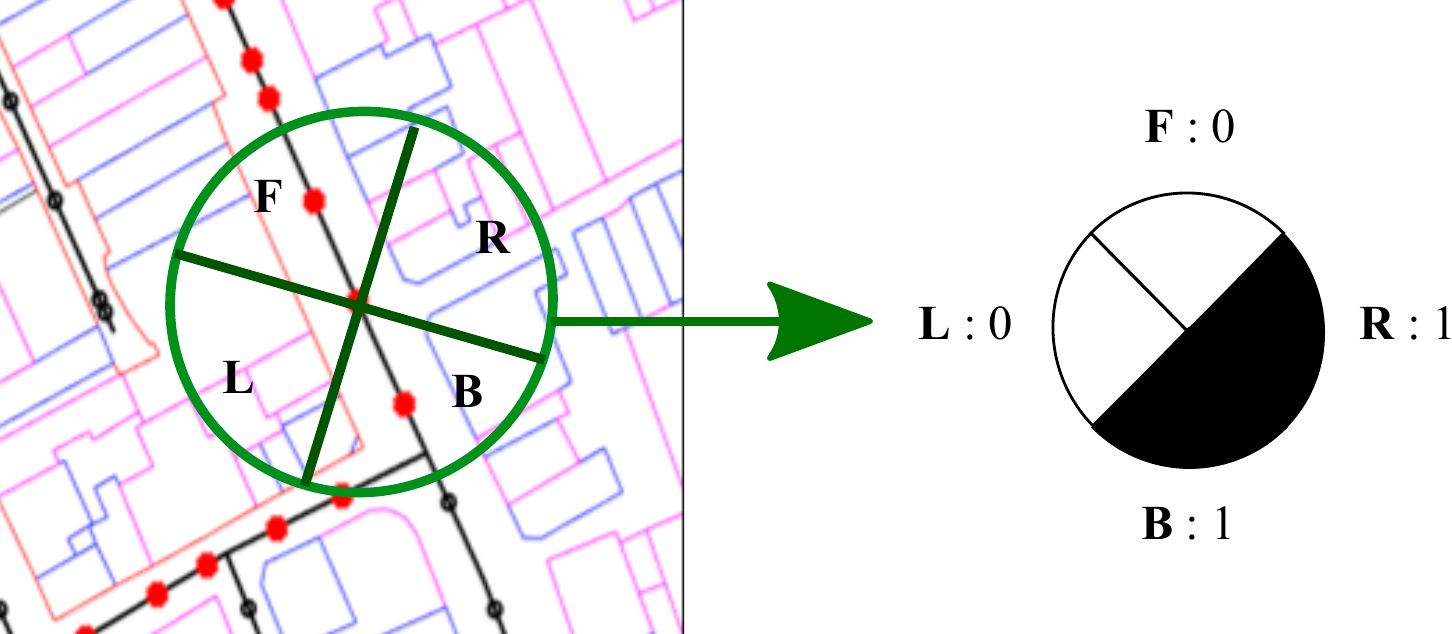}
\end{center}
\vspace*{-3ex}
\caption{Generation of a BSD from the vector map.}
\label{fig:bsdgen}
\vspace*{-3ex}
\end{figure}

We denote the finite set of locations in an area of interest by $\mathcal{L}=\{l_1, l_2, \ldots, l_N\}$, where $N$ is the total number of locations. Associated with each location $l_i$ is a BSD, which we denote by the binary string $d_i$, with $d_{ij}$ denoting the $j$th bit, and define $\mathcal{D}=\{d_1, d_2, \ldots, d_N\}$ as the set of all descriptors. In this work, $j\in\{1,2,3,4\}$ and each bit of a BSD denotes the presence or not of a junction or a gap between buildings in one of four viewing directions centred on location $i$.  These are derived from the vector map as follows
\be
d_{ij}=\left \{ \begin{array}{ll}
	 JUNC(V_{ij}) & \mbox{if $j\in\{1,2\}$}\\
	 BGAP(V_{ij}) & \mbox{if $j\in\{3,4\}$}
	 \end{array} \right .
	 \label{eqn:bsdgen}
\ee
where ($V_{i1}$,$V_{i2}$) and ($V_{i3}$,$V_{i4}$) denote the (front,back) and (left,right) viewing directions at location $i$, respectively. The functions $JUNC(V_{ij})$ and $BGAP(V_{ij})$ return 1 if there exists a junction or a gap between buildings, respectively, in direction $V_{ij}$, and 0 otherwise. As illustrated in Figure \ref{fig:bsdgen}, a feature is deemed to be present in a viewing direction if one lies within the relevant quadrant of a circle of a given radius centred on the location of interest, where the front and back viewing directions are aligned with that of the road upon which the location sits. In the experiments we set the viewing distance radius to be 30m, which is similar to that used in \cite{Seff2016}.

For localisation we need to estimate a BSD for a location from images captured in each of the four viewing directions. We do this using binary classifiers, trained to detect the presence or not of the relevant semantic feature. Given image $I_{ij}$ at location $i$ in viewing direction $V_{ij}$, the estimated BSD is given by
\be
\hat{d}_{ij}=\left \{ \begin{array}{ll}
	 DETECT_{JUNC}(I_{ij}) & \mbox{if $j\in\{1,2\}$}\\
	 DETECT_{BGAP}(I_{ij}) & \mbox{if $j\in\{3,4\}$}
	 \end{array} \right .
	 \label{eqn:bsdest}
\ee
where $DETECT_{JUNC}(I_{ij})$ and $DETECT_{BGAP}(I_{ij})$ return 1 if a junction or a gap, respectively, are detected in image $I_{ij}$, and 0 otherwise, i.e. they mirror the BSD generation functions in Equation \ref{eqn:bsdgen}.

We use a CNN approach to design the binary classifiers $DETECT_{JUNC}$ and $DETECT_{BGAP}$. For training data, we make use of the correspondence between the vector maps in OSM and the images in GSV in a similar manner to that used in \cite{Seff2016}. For each feature type - junctions and gaps between buildings - we collect positive samples by identifying the locations of the relevant features in OSM and storing the images from the corresponding locations (based on latitude and longitude) and relevant viewing directions from GSV, ensuring that we get a uniform mix of viewing scenarios. For example, in the case of junctions, we use front and back facing images aligned with the road and ensure that we have examples that cover the range of distances from the junction up to the viewing radius used in the generation of the BSDs. The training set is then completed by collecting approximately the same of number of negative samples in the corresponding viewing directions but not containing the feature of interest. In the experiments, we used a training sets consisting of 440,000 images per classifier taken from 220,000 locations in 23 different cities in the UK. None of these locations were used for evaluating the classifiers or in the localisation experiments.

\begin{figure}[t]
\vspace*{1ex}
\centerline{
\begin{tabular}{cc}
\includegraphics[width=0.22\columnwidth]{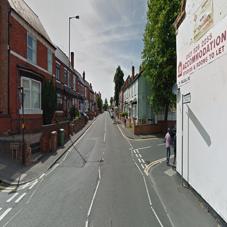}
\includegraphics[width=0.22\columnwidth]{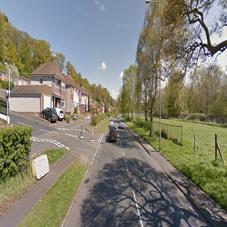}&
\includegraphics[width=0.22\columnwidth]{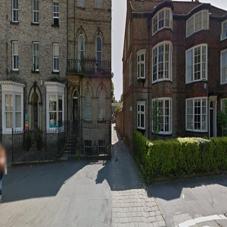}
\includegraphics[width=0.22\columnwidth]{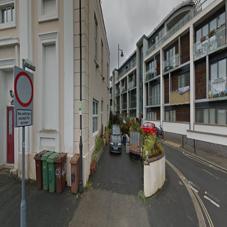}\\
\includegraphics[width=0.22\columnwidth]{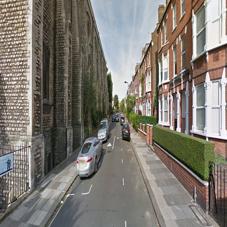}
\includegraphics[width=0.22\columnwidth]{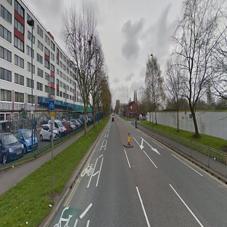}&
\includegraphics[width=0.22\columnwidth]{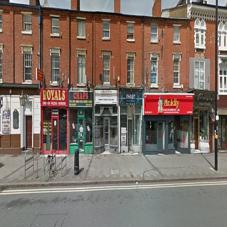}
\includegraphics[width=0.22\columnwidth]{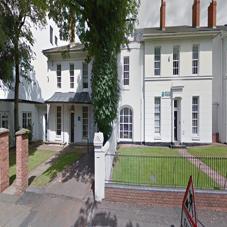}\\
(a) & (b)
\end{tabular}
}
\vspace*{-1ex}
\caption{Examples of positive (feature present) and negative (feature not present) images from the training datasets used for the semantic classifiers: (a) junction (top) and no junction (bottom); (b)  gap (top) and no gap (bottom).}
\label{fig:training}
\vspace*{-3ex}
\end{figure}

We implemented the classifiers by using our training dataset to fine-tune an off-the-shelf pre-trained CNN. Specifically, we started from the pre-trained Places205-AlexNet model \cite{Zhou2014}, designed for scene classification in urban environments, which aligns with our application, and derived from the pre-trained AlexNet model \cite{Krizhevsky2012}.We used colour images cropped from GSV panoramas in the required viewing direction corresponding to a $90^\circ$ horizontal field of view and resized to $227\times 227$ pixels. The latter results in some distortion but given that we used the same process for both training and testing, this was not deemed to be an issue. Examples of positive and negative images from the training dataset are shown in Figure \ref{fig:training}. We tested performance of each classifier using two test sets of 8000 images taken from the same 23 cities but at locations not within the training set and with an equal number of positives and negatives samples, i.e. feature present and not present.

\begin{figure}[t]
\vspace*{1ex}
\centerline{
\begin{tabular}{cc}
\includegraphics[width=0.22\columnwidth]{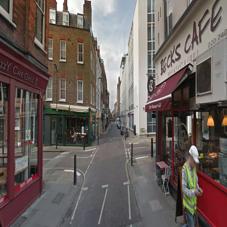}
\includegraphics[width=0.22\columnwidth]{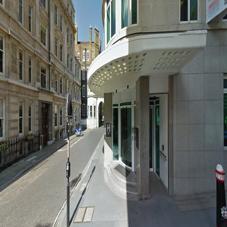}&
\includegraphics[width=0.22\columnwidth]{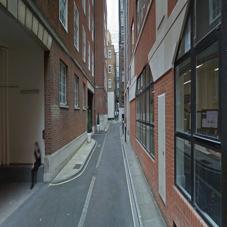}
\includegraphics[width=0.22\columnwidth]{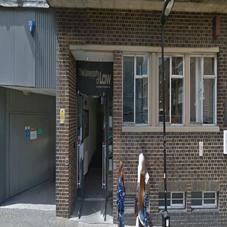}\\
\includegraphics[width=0.22\columnwidth]{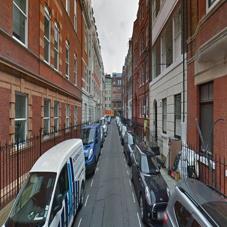}
\includegraphics[width=0.22\columnwidth]{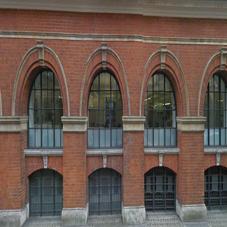}&
\includegraphics[width=0.22\columnwidth]{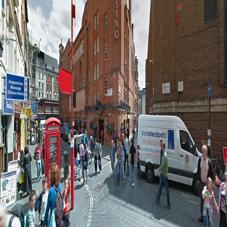}
\includegraphics[width=0.22\columnwidth]{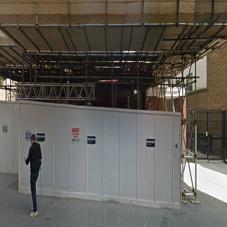}\\
(a) & (b)
\end{tabular}
}
\vspace*{-1ex}
\caption{Examples of semantic classifications: (a) true positives (top) and true negatives (bottom);  (b) false positives (top) and false negatives (bottom). In both (a) and (b) examples are arranged as: junction (top-left); gap (top-right); no junction (bottom-left); no gap (bottom-right).}
\label{fig:classresults}
\vspace*{-3ex}
\end{figure}

Both classifiers gave good balanced performance in detecting the presence and non-presence of junctions and gaps, with precision and recall values of $0.75\pm 0.02$ on the test set.  Examples of correct classifications (true positives and true negatives) and incorrect classifications (false positives and false negatives) are shown in Figure \ref{fig:classresults}. Note that the latter illustrate the difficulty of the task. For example, the bottom left view in Figure \ref{fig:classresults}b contains a junction which is significantly obscured and was incorrectly classified as containing no junction, whilst the 2-D map indicates that the bottom right view should contain a gap, but the site appears to be under redevelopment and has been incorrectly classified as not  containing a gap. The latter is an example of inaccuracies within the OSM data. 

\section{ROUTE DESCRIPTORS AND TURN PATTERNS}

As noted earlier and as we demonstrate later, on their own the above binary descriptors are not sufficiently discriminative to  identify a location uniquely and allow localisation. This is true even if we were able to design perfect classifiers for extracting the descriptors from images. The simplicity of the representation, whilst being extremely compact, means that there are many locations with similar descriptors. We address this ambiguity in two ways. First, we concatenate descriptors along routes corresponding to adjacent locations, constructing {\em route descriptors}, which prove to be highly discriminative once the routes reach a certain length. Once this length is reached, then localisation can proceed at the rate that new locations are visited, i.e. enabling tracking, by matching with a database of all possible route descriptors constructed offline. Secondly, we introduce further disambiguation by incorporating {\em turn patterns} observed along routes into the representation, i.e. the sequence no turn and turn (left or right) at each location along a route, and using these to identify the most likely match within the database.

Let $A$ be an $N\times N$ adjacency matrix, such that $A_{ij}=1$ if locations $l_i$ and $l_j$ are adjacent, and $A_{ij}=0$ otherwise. Locations are regarded as adjacent if on the 2-D map they are connected by a road and there are no other locations between them. A {\em route} is then defined as a finite sequence of adjacent locations, i.e. the route $r=(l_{\gamma(1)}, l_{\gamma(2)},\ldots,l_{\gamma(N_r)})$ is of length $N_r$, where $\gamma(i)$ defines a sequence of adjacent locations such that $A_{\gamma(i)\gamma(i+1)}=1$, $\forall \;1\leq i< N_r$. For simplicity we have restricted ourselves to routes that do not loop or turn back on themselves, i.e. $\gamma(i)\neq\gamma(j)$, $\forall$ $i\neq j$, $1\leq i,j\leq N_r$, but the method could be readily extended to deal with such cases. We define ${\mathcal R}_M$ as the set of all such routes up to length $M$ defined amongst all the locations in ${\mathcal L}$. Associated with each route is a {\em route descriptor}, consisting of the sequence of BSDs corresponding to the locations along the route, i.e. $s=(d_{\gamma(1)}, d_{\gamma(2)},\ldots,d_{\gamma(N_r)})$, and we define $\mathcal{S}_M$ as the set of all route descriptors corresponding to the routes in $\mathcal{R}_M$. 

To incorporate turn information into the representation, we define a binary {\em turn pattern} $t=(t_{\gamma(1)}, t_{\gamma(2)},\ldots,t_{\gamma(N_r-1)})$ associated with a route $r$. The $i$th bit of $t$ indicates whether a left and right turn is present between locations $l_{\gamma(i)}$ and $l_{\gamma(i+1)}$, i.e. $t_{\gamma(i)}=TU\!RN(\theta_{\gamma(i)},\theta_{\gamma(i+1)})$, where $\theta_{\gamma(i)}$ denotes the front facing direction at location $l_{\gamma(i)}$ and 
\be
TU\!RN(\theta_i,\theta_j)=\left \{ \begin{array}{ll}
	 1 & \mbox{if $\lfloor \theta_i-\theta_j\rfloor\geq \tau$}\\
	 0 & \mbox{otherwise}
	 \end{array} \right .
	 \label{eqn:turndef}
\ee
where $\lfloor \theta_i-\theta_j\rfloor$ denotes the absolute value of the smallest angle between $\theta_i$ and $\theta_j$, and $\tau$ is an angle threshold, which we set to be $60^\circ$ to ensure that we only include significant turns. Thus $t$ represents the sequence of turns that take place along a route. We define $\mathcal{T}_{M}$ to be the set of such turn patterns  corresponding to the routes in $\mathcal{R}_M$.

\section{LOCALISATION AND BOOTSTRAPPING \label{sec:localisation}}

Consider an autonomous system making its way through an urban environment, moving between locations in $\mathcal{L}$ along a specific route of length $<M$. At any given location, our goal is to identify its current location by recognising the route taken to date, consisting of the current location plus the previous $N_r-1$ locations, say. We do this by comparing its estimated route descriptor $\hat{s}$ (obtained by concatenating the estimated BSDs at each location) with those in  $\mathcal{S}_{N_r}\subset \mathcal{S}_M$ and its turn pattern $\hat{t}$ with those in $\mathcal{T}_{N_r}\subset \mathcal{T}_{M}$, hence determining the most likely route from those in $\mathcal{R}_{N_r}\subset \mathcal{R}_M$.

It is important to note that in this work we assume that there is a one-to-one correspondence between the locations in our 2-D map and the locations in the environment. This enables us to do a direct comparison between estimated route descriptors and those in the database. When using GSV and OSM data this can be ensured by selecting OSM locations corresponding to the known locations in GSV. In a practical system, we would need a method of forming such one-to-one correspondence or alternatively, a means overcoming the lack of it. We discuss this further in Section \ref{sec:conclusions}.

We define the most likely route $r^*\in{\cal R}_{N_r}$ as being the route whose route descriptor $s^*$ is closest to $\hat{s}$ and whose turn pattern $t^*$ matches $\hat{t}$, i.e. such that
 \be
 s^*=\argmin_{s\in {\mathcal{S}_{N_r}}} DIST(s,\hat{s})
 \label{eqn:closestroute}
 \ee 
 and
 \be
 DIST(t^*,\hat{t})=0
 \ee
 where $DIST(a,b)$ denotes the Hamming distance between two binary strings $a$ and $b$. For long routes ($N_r>20$, say) the number of elements in $\mathcal{S}_{N_r}$ becomes very large ($>500\times 10^3$, rising to near $40\times 10^6$ for $N_r=40$), and thus we use an efficient pattern matching algorithm based on a BK-tree \cite{Burkhard1973} to find the closest route descriptor in Equation (\ref{eqn:closestroute}).

Note from the above that we assume that the turn pattern for the query route is correct, but allow errors in the estimate of the route descriptor due to the non-perfect classifiers used in the semantic feature detectors. Our motivation for the former is that in practice detecting significant left or right turns by an autonomous system can be achieved reliably and thus requiring an exact match is reasonable. Note, however, that as we show later, turn patterns alone are not sufficient to achieve localisation, as many routes share the same turn pattern, and it is their combination with route descriptors that gives the required level of distinctiveness.

The above provides an indication of the most likely location given the current route. However it gives no indication as to the confidence in the estimate. There are a number of possibilities for this, including basing it on the distance between $s^*$ and $\hat{s}$ and/or the distance of $s^*$ from the second best matching route descriptor. We found that a consistency metric proved to be most effective, in which we deem a route to be localised if there is sufficient overlap between the most likely routes $r^*$ for a number of successive locations. We set the overlap to be 80\% of the locations need to be the same and we required this to occur for 5 successive locations. In essence, if successive query routes are matching with routes that have significant overlap then it is a good indicator that successful localisation has been achieved. 

We also demonstrate later that once the above consistency criterion has been met, the query route length can be fixed and localisation proceed by successively updating the query route by appending the latest BSD onto the end and removing the first descriptor. Thus, the phase during which the query route grows can be regarded as a {\em bootstrapping process}, during which the route descriptor extends until it becomes sufficiently distinct to allow localisation. Once complete, then {\em continuous tracking} can take place using the fixed length query at the same rate as the BSD are created at successive locations. An example of bootstrapping and tracking is shown in the video submitted as supplementary material.
 
\section{PROBABILISTIC FORMULATION}

The above localisation process can also be considered in probabilistic terms. Given an estimated BSD $\hat{d}$ obtained at a single location $l$, say, then the conditional probability that $l$ corresponds to $l_i\in \mathcal{L}$ can be written as
\be
P(l_i | \hat{d})=P(l_i | d_i)P(d_i | \hat{d})\propto P(l_i|d_i)P(\hat{d}|d_i)
\label{eqn:locprob}
\ee
where we assume that all descriptors $d_i$ are equally likely. Note that the term $P(l_i|d_i)$ expresses the uniqueness of the ground-truth descriptor $d_i$ derived from the 2-D map. Since our descriptors are only 4-bits long, then for a large number of locations, e.g. 6000 in the experiments, $P(l_i|d_i)<<1$, indicating that many locations have the same descriptors and hence localisation is not possible. Given that we have an estimate of the accuracy of our classifiers and hence the detectors $DETECT_{JUNC}$ and $DETECT_{BGAP}$, we can approximate the likelihood $P(\hat{d}|d_i)$ in terms of the Hamming distance $h$ between $d_i$ and $\hat{d}$, i.e.
\be
P(\hat{d}|d_i)\propto q^{4-h}(1-q)^h
\label{eqn:likelihood}
\ee
where $q$ is the probability of correctly detecting the presence or not of both junctions and gaps (we assume the same value for both probabilities for simplicity, but as noted in Section \ref{sec:bsd}, we also observed similar values in practice of $\approx 0.75$). 

Extending the above to routes, we obtain the following conditional probability that the route descriptor estimate $\hat{s}=(\hat{d}_1,\hat{d}_2,\ldots,\hat{d}_{N_r})$ corresponds to route $r\in \mathcal{R}_{N_r}$
\be
P(r|\hat{s}) = P(r|s)P(s|\hat{s})=P(r|s)P(\hat{s}|s)
\ee
Hence from Equations (\ref{eqn:locprob}) and (\ref{eqn:likelihood}) and assuming independence between descriptors
\begin{eqnarray}
P(r|\hat{s}) &\propto& P(r|s)\prod_{i=1}^{N_r}P(\hat{d}_i|d_{\gamma(i)})\\
& \propto & P(r|s)q^{4N_r-H}(1-q)^{H}
\end{eqnarray}
where $H$ denotes the Hamming distance between $s$ and $\hat{s}$. Here, $P(r|s)$ expresses the uniqueness of the route descriptor $s$, which as we demonstrate later is high for a sufficiently long routes and thus $P(r|s)\rightarrow 1$, giving 
\be
P(r|\hat{s})\propto q^{4N_r-H}(1-q)^{H}
\ee
Using this expression we can obtain an estimate of the likelihood ratio of one route $r_i$ over another $r_j$ for a given $\hat{s}$
\be
\frac{P(r_i|\hat{s})}{P(r_j|\hat{s})}=\frac{(1-q)^{H_i-H_j}}{q^{H_i-H_j}}
\ee
where $H_i$ is the Hamming distance between $s_i$ and $\hat{s}$. Hence for $q=0.75$, this gives a likelihood ratio of $1/3^\delta$ for a difference $\delta$ in Hamming distance from the estimated route descriptor, which is significant. For example,  a route whose descriptor is $\delta$ bits closer in Hamming distance to the estimated descriptor, is $3^\delta$ times more likely to be the correct route. Thus, as we demonstrate in the next section,  even with a detector accuracy of only 75\% for individual BSDs, the concatenation of descriptors along routes can lead to a high degree of distinctiveness for long enough routes.

\begin{figure}[t]
\begin{center}
\includegraphics[width=0.95\columnwidth]{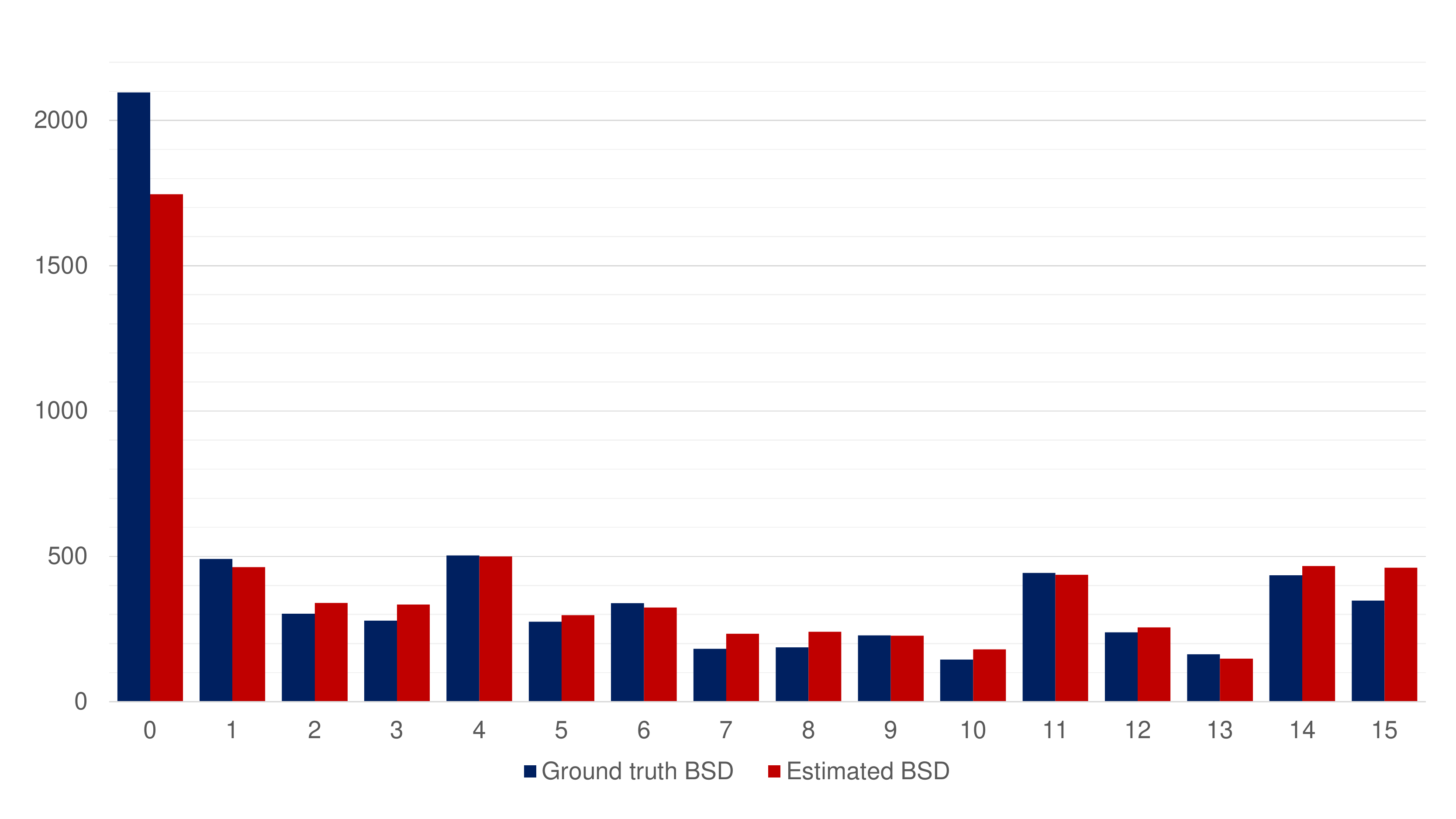}
\end{center}
\vspace*{-3ex}
\caption{Histogram showing the distribution of 4-bit ground-truth (blue) and estimated (red) BSDs obtained from OSM and GSV images, respectively.}
\label{fig:bsddist}
\vspace*{-3ex}
\end{figure}

\section{EXPERIMENTS \label{sec:expts}}

We evaluated the performance of the method using GSV and OSM data for a $2.5$ $\mbox{km}^2$ region of London. None of the locations within the region were used to train the semantic classifiers. The region consisted of 6656 GSV locations and from each location we gathered images corresponding to the four viewing directions, from which we estimated 4-bit BSDs using the classifiers described in Section \ref{sec:bsd}.

\begin{figure}[t]
\vspace{2ex}
\begin{center}
\includegraphics[width=\columnwidth]{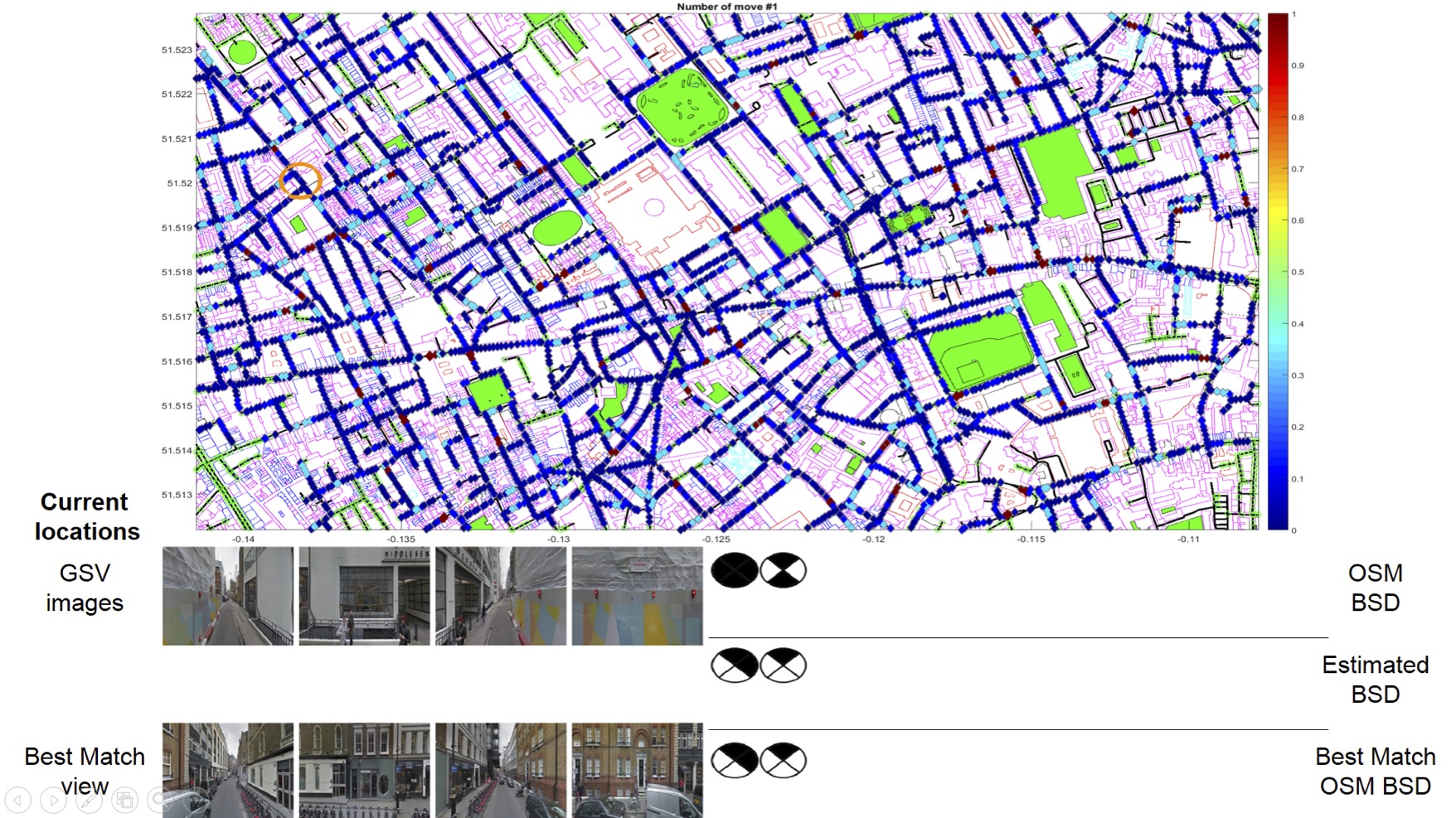}\\[2ex]
\includegraphics[width=\columnwidth]{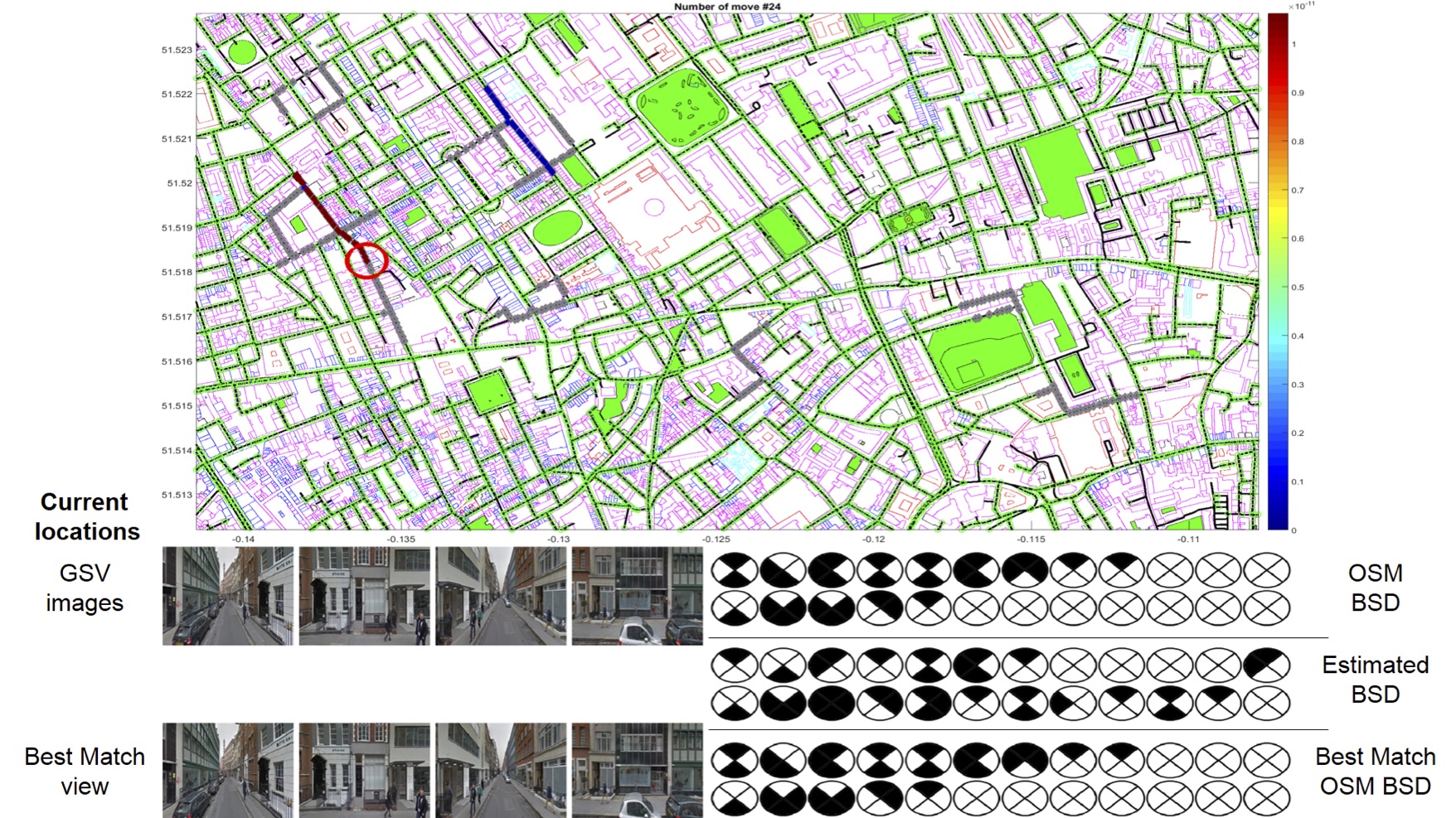}
\end{center}
\vspace*{-2ex}
\caption{Snapshots of the localisation process for test route lengths of 2 (top) and 24 locations (bottom). See text for explanation.}
\label{fig:videoframes}
\vspace*{-3ex}
\end{figure}

To illustrate the distribution of descriptors across the region and the performance of the classifiers, Figure \ref{fig:bsddist} shows the histogram of 4-bit ground-truth descriptors (obtained from OSM, shown in blue) and estimated descriptors, shown in red, where the horizontal axis corresponds to the 16 possible 4-bit patterns. The predominance of BSDs with pattern `0000' corresponding to locations in which there is neither gaps between buildings to the left or right, nor junctions towards the front or back, results from the fact that many locations between junctions have these characteristics as can be seen from the 2-D map of the area shown in Figure \ref{fig:videoframes}. Note that the distribution of the estimated descriptors is close to that of the ground-truth due to the accuracy of the classifiers, i.e. approximately 75\%. 

\begin{figure}[t]
\begin{center}
\includegraphics[width=0.95\columnwidth]{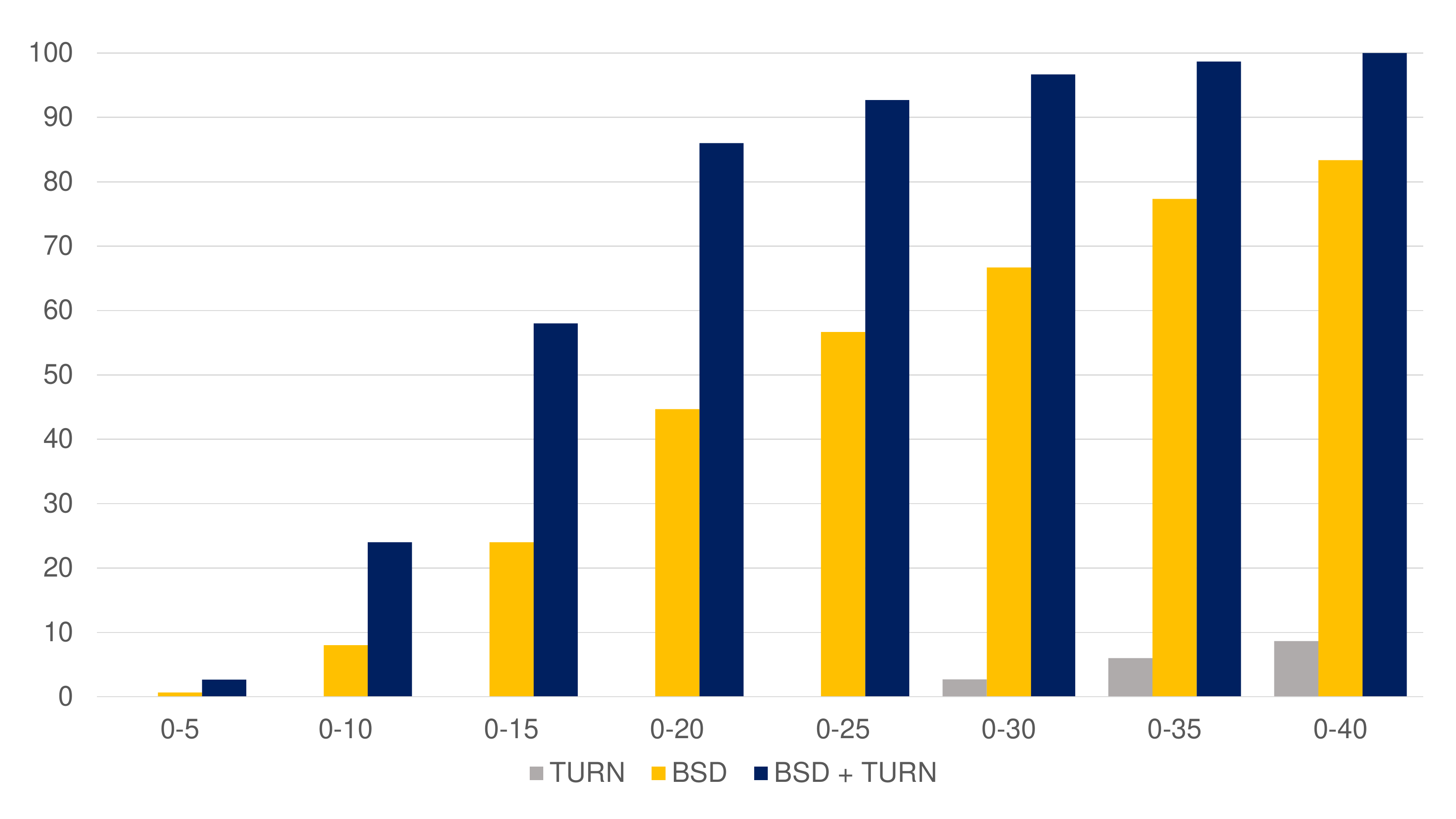}
\end{center}
\vspace*{-3ex}
\caption{Accuracy of localisation (\% of correctly identified routes) versus route length using turn patterns (grey), route descriptors (yellow), and route descriptors with turn patterns (blue).}
\label{fig:localacc}
\vspace*{-3ex}
\end{figure}

To assess the performance of the route based localisation, we considered route lengths up to a maximum of $M=40$ locations. We tested the method using 150 test routes and for each we recorded the route length at which localisation was achieved according to the consistency criterion, i.e. 5 successive consistent localisations. The results are shown in Figure \ref{fig:localacc}, which shows the percentage of routes that were correctly localised within route lengths of 0-5, 0-10, ..., 0-40 locations. We have shown the results for three methods of matching routes: using only turn patterns (grey); using only route BSDs  (yellow); and using both BSDs and turn patterns (blue). Note that the latter outperforms the others by a significant margin and that BSDs alone also significantly outperform turn patterns, which only manage to localise $<10\%$ of routes even with a route length of 40 locations. This clearly demonstrates the potential of the BSD approach. Note in particular that over 85\% of routes are correctly localised even when only using routes consisting of up to 20 locations, which corresponds to approximately 200 meters in GSV.

\begin{figure}[t]
\vspace{2ex}
\begin{center}
\includegraphics[width=0.95\columnwidth]{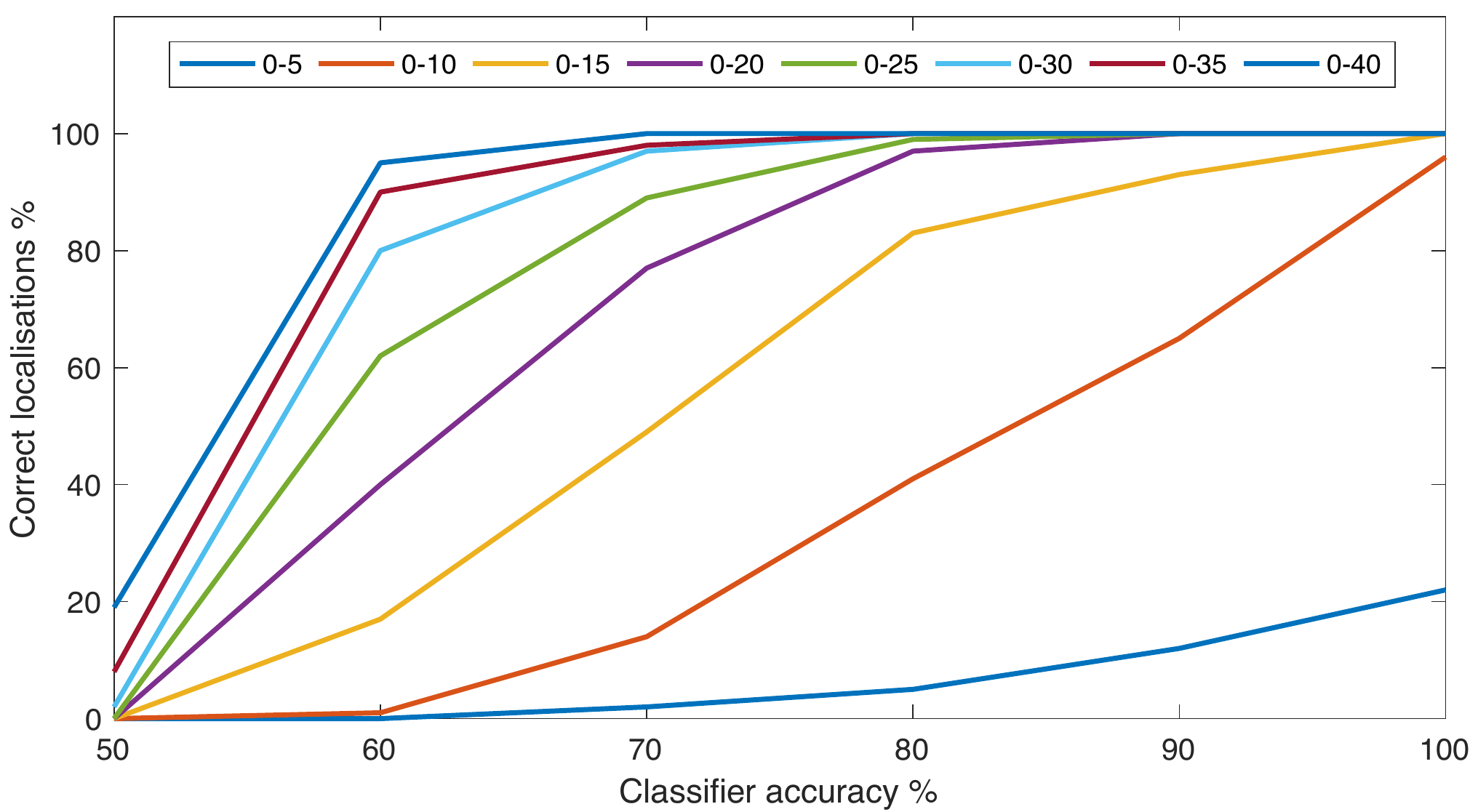}
\end{center}
\vspace*{-3ex}
\caption{Accuracy of localisation (\% of correctly identified routes) versus classifier accuracy for different ranges of route length.}
\label{fig:classacc_vs_localisations}
\vspace*{-3ex}
\end{figure}

It is also interesting to consider the significance of the classifier accuracy. Given that we know the ground-truth BSDs, we investigated using BSDs 'estimated' using classifiers with different accuracy (we assumed the same accuracy for both detecting the presence or not of junction and gaps). A plot of the percentage of correctly localised routes versus the accuracy of the classifiers is shown in Figure \ref{fig:classacc_vs_localisations} for the different route length ranges used in Figure \ref{fig:localacc}. Thus that at 75\% accuracy, which we obtained from our trained classifiers, over 85\% of routes are predicted to be correctly localised within 0-20 locations, which agrees with our findings in Figure \ref{fig:localacc}. Note also that if classifier accuracy were increased to beyond 80\%, then 80-90\% of routes could be correctly localised using $<15$ locations, which again illustrates the potential of the BSD approach.

\begin{figure}[t]
\vspace{0ex}
\begin{center}
\includegraphics[width=0.9\columnwidth]{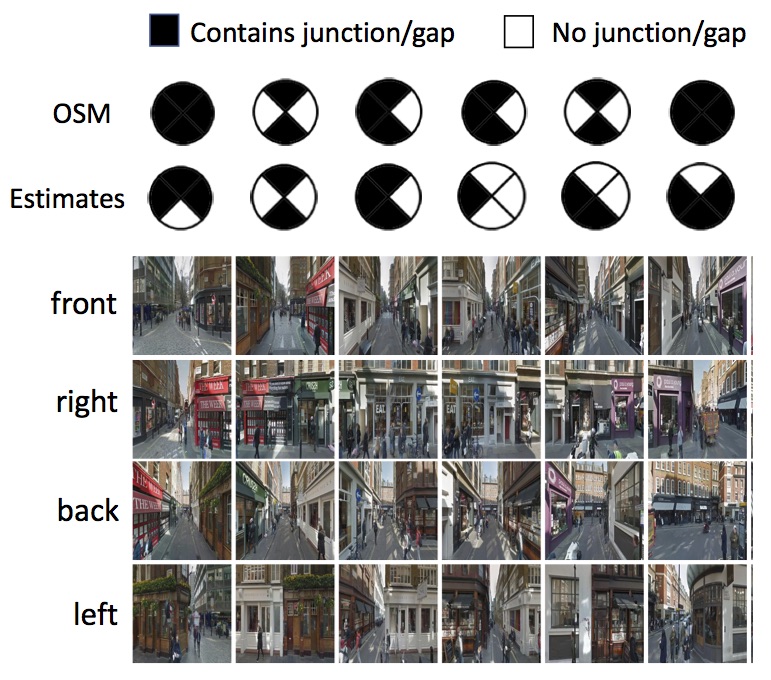}
\end{center}
\vspace*{-3ex}
\caption{Examples of BSD ground-truths, from OSM, and BSD estimates, from the classification of the GSV images in four directions as shown.}
\label{fig:exbsdests}
\vspace*{-3ex}
\end{figure}
 
\begin{figure}[t]
\begin{center}
\vspace{2ex}
\includegraphics[width=0.95\columnwidth]{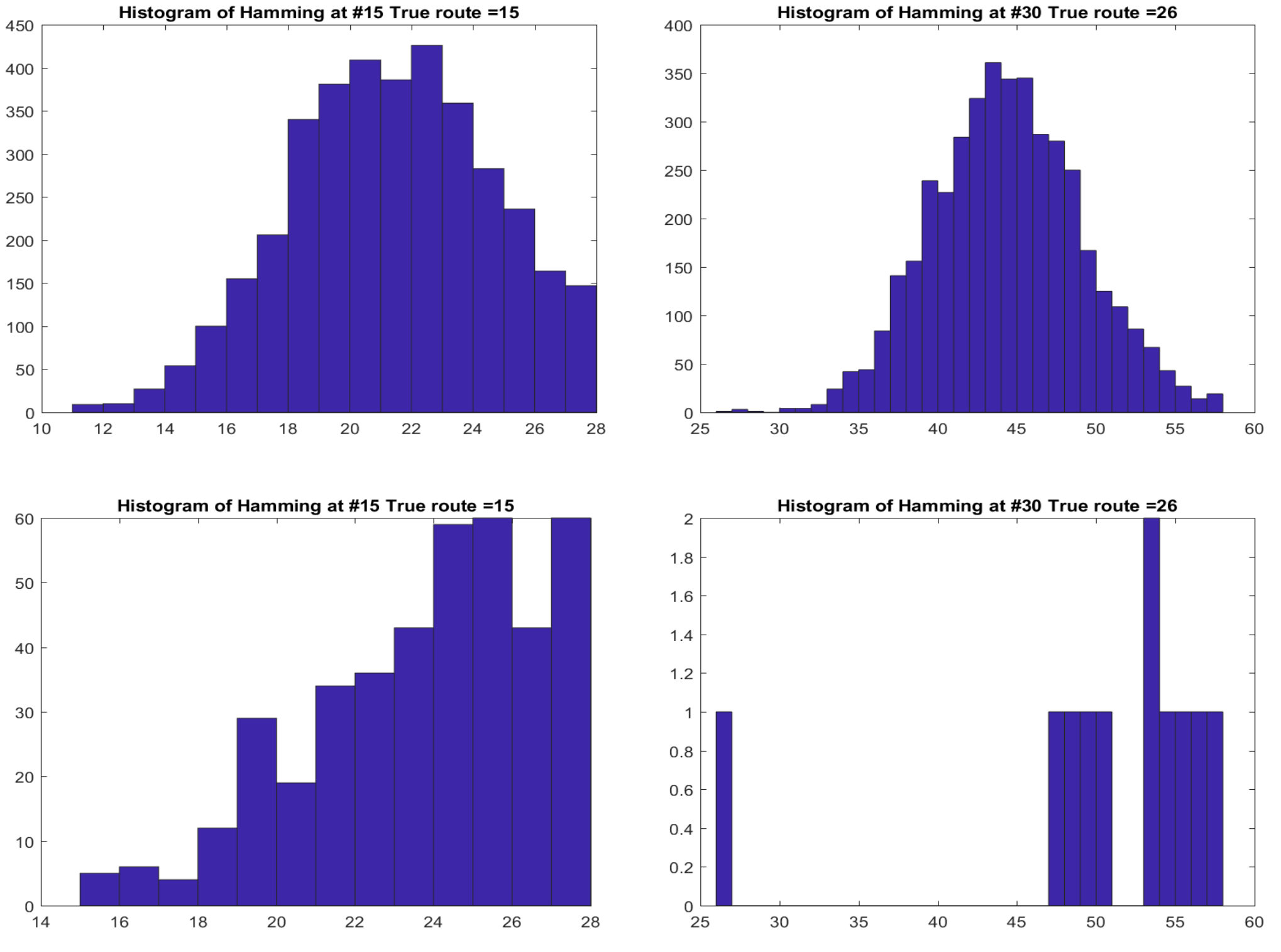}
\end{center}
\vspace*{-3ex}
\caption{Histograms of Hamming distances between a test route descriptor and those in the database for route lengths of 15 (left) and 30 (right) locations, with (bottom) and without (top) using turn patterns.}
\label{fig:hammingdists}
\vspace*{-3ex}
\end{figure}

Examples of estimated BSDs, their corresponding images in the four viewing directions and the ground-truth BSDs from OSM for part of a route are shown in Figure \ref{fig:exbsdests}. Note the deviation of the BSD estimates from the ground-truth, which results from the inaccuracy of the classifiers. The challenging nature of the detection task can be seen from the images. This confirms the utility of concatenating BSDs along a route in order to gain uniqueness and hence enable localisation. Figure \ref{fig:hammingdists} further illustrates this, which shows the distribution of Hamming distances from descriptors in the database for a given test (query) route at lengths of 15 (left) and 30 (right) locations, with and without  using turn patterns (bottom and top, respectively). The correct matches for lengths 15 and 30 have Hamming distances of 15 and 26, respectively.  When the test route length is 15 locations, the correct route is not the closest (there are other Hamming distances with values $<15$), although using turns (bottom) significantly reduces the number of routes close to the query route (note that the vertical axes in Figure \ref{fig:hammingdists} have significantly different ranges). With 30 locations and without using turns, the correct route does become equal closest with 18 others and there are a significant number of others close by. In contrast, using turn patterns with 30 locations drastically reduces the number of candidate routes and the correct route becomes the closest by a Hamming distance margin of over 20.

To illustrate the localisation process, Figure \ref{fig:videoframes} shows snapshots of the localisation of a test route at route lengths of 2 and 24 locations. It shows the OSM 2-D map and the locations are indicated by coloured squares along roads, where the colour indicates the closeness between their route descriptor and that of the test route (where locations share routes, then the closest descriptor difference is shown). The latest location along the test route is indicated by the orange/red circle, where orange indicates that the route has yet to be correctly and consistently localised, and red indicates that localisation has been achieved. The BSDs, estimated and ground-truths, along with their images, are shown below the 2-D maps. Note that the bottom row of images show the views at the closest (best) match locations, but are not used in the matching process. With route length of 2, the majority of locations have a low likelihood of being correct (dark blue), whilst a small number of disparate locations have a high likelihood (dark red). This reflects the lack of distinctiveness of two 4-bit BSDs. In contrast, once  24 locations are reached, the route has been successfully localised and the vast majority of other locations/routes have been eliminated (their squares are not shown), reflecting the confidence of the localisation. A video showing the complete process has been submitted as supplementary material.

\section{CONCLUSIONS \label{sec:conclusions}}

We have presented a novel approach to position localisation in urban areas and to the best of our knowledge it is the first example of linking 2-D maps to images over large areas. The key contribution is the demonstration that compact binary semantic descriptors concatenated over routes are sufficiently distinctive to enable localisation and that the representation is vastly smaller than that used in image to image database approaches. Moreover, the use of simple semantic classification offers the potential for invariance to changing environment conditions, which is something that we wish to demonstrate in future. In addition, the reported work relies on an assumption of one-to-one correspondence between map and image locations, achieved by using OSM and GSV data, and this needs to be addressed for developing a practical system, which we are in the process of doing.

\bibliographystyle{IEEEtran}
\bibliography{iros18}

\end{document}